\documentclass{article}

\usepackage{microtype}
\usepackage{graphicx}
\usepackage{subcaption}
\usepackage{booktabs} 
\usepackage{amsmath}
\DeclareMathOperator*{\argmax}{arg\,max}
\DeclareMathOperator*{\argmin}{arg\,min}
\usepackage[draft]{hyperref}

\usepackage[arxiv]{icml2020}

\usepackage{multirow}
\usepackage{cleveref}
\usepackage[table]{xcolor}

\usepackage[
  separate-uncertainty = true,
  multi-part-units = repeat
]{siunitx}

\usepackage{bm}
\usepackage{mathtools}
\usepackage{graphicx}
\usepackage{amsfonts}

\icmltitlerunning{
Probabilistic Dual Network Architecture Search on Graphs}

\newcommand{\mymatrix}[1]{\bm{\mathit{#1}}}

\newcommand*\bigcdot{\mathpalette\bigcdot@{.5}}
\newcommand{\eg}{\textit{e.g\@.}}
\newcommand{\etc}{\textit{etc.}}

\newcommand{\ie}{\textit{i.e\@.}}

\newcommand{\realset}{\mathbb{R}}

\begin{document}

\twocolumn[
\icmltitle{Probabilistic Dual Network Architecture Search on Graphs}



\icmlsetsymbol{equal}{*}

\begin{icmlauthorlist}
\icmlauthor{Yiren Zhao}{equal,cam}
\icmlauthor{Duo Wang}{equal,cam}
\icmlauthor{Xitong Gao}{equal,siat}
\icmlauthor{Robert Mullins}{cam}
\icmlauthor{Pietro Lio}{cam}
\icmlauthor{Mateja Jamnik}{cam}
\end{icmlauthorlist}

\icmlaffiliation{cam}{University of Cambridge, Cambridge, UK}
\icmlaffiliation{siat}{Shenzhen Institue of Advanced Technology, Shenzhen, China}

\icmlcorrespondingauthor{Yiren Zhao}{yaz21@cam.ac.uk}

\icmlkeywords{Machine Learning, ICML}

\vskip 0.3in
]



\printAffiliationsAndNotice{\icmlEqualContribution} 

\begin{abstract}

We present the first differentiable Network Architecture Search (NAS) for Graph Neural Networks (GNNs). GNNs show promising performance on a wide range of tasks, but require a large amount of architecture engineering. First, graphs are inherently a non-Euclidean and sophisticated data structure, leading to poor adaptivity of GNN architectures across different datasets. Second, a typical graph block contains numerous different components, such as aggregation and attention, generating a large combinatorial search space. To counter these problems, we propose a Probabilistic Dual Network Architecture Search (PDNAS) framework for GNNs. PDNAS not only optimises the operations within a single graph block (micro-architecture), but also considers how these blocks should be connected to each other (macro-architecture). The dual architecture (micro- and marco-architectures) optimisation allows PDNAS to find deeper GNNs on diverse datasets with better performance compared to other graph NAS methods. Moreover, we use a fully gradient-based search approach to update architectural parameters, making it the first differentiable graph NAS method. PDNAS outperforms existing hand-designed GNNs and NAS results, for example, on the PPI dataset, PDNAS beats its best competitors by $1.67$ and $0.17$ in F1 scores.

\end{abstract}

\section{Introduction}

Graphs are a ubiquitous structure and are widely used
in real-life problems,
\eg~computational biology \cite{zitnik2017predicting},
social networks \cite{hamilton2017inductive},
knowledge graphs \cite{lin2015learning}, \etc{}
Graph Neural Networks (GNNs) follow a message passing
(node aggregation) scheme to gradually propagate information
from adjacent nodes at every layer.
However, due to the varieties of non-Euclidean data structures, GNNs tend to be
less adaptive than traditional convolutional neural networks,
thus it is common to re-tune the network architecture for each new dataset.
For instance, GraphSage \cite{hamilton2017inductive}
shows networks are sensitive to the number of
hidden units on different datasets;
jumping knowledge networks
demonstrate that the optimal concatenation strategy between layers
varies for different datasets \cite{xu2018representation}.
Furthermore, the challenge of designing a new GNN architecture
typically involves a considerably larger design space.
A single graph block normally comprises multiple connecting sub-blocks,
such as linear layers, aggregation, attention, \etc,
each sub-block can have multiple candidate operations,
and thus provides a large combinatorial architecture search space.
The formidable search space and the lack
of transferability of GNN architectures
present a great challenge
in deploying GNNs rapidly to various real-life scenarios.

Recent advances in neural network architecture search (NAS) methods
show promising results on convolutional neural networks
and recurrent neural networks
\cite{zoph2016neural,liu2018darts,casale2019probabilistic}.
NAS methods are also applicable to graph data, recent work
uses NAS based on reinforcement learning (RL) for GNNs
and achieves state-of-the-art accuracy results
\cite{gao2019graphnas,zhou2019auto}.
RL-based NAS, however, has the following shortcomings.
First, RL requires a full train and evaluate cycle for each architecture
that is considered;
making it computational expensive \cite{casale2019probabilistic}.
Second, existing GNN search methods focus only on the micro-architecture.
For instance, only the activation function,
aggregation method, hidden unit size,
\etc~of each graph convolutional block are considered in the search.
However, it has been observed that performance can be improved if
shortcut connections, similar to residual connections in
CNNs \cite{he2016deep}, are added to adapt neighborhood ranges
for a better structure-aware representation
\cite{xu2018representation}.
This macro-architecture configuration
of how blocks connect to each other via shortcuts
is not considered in previous Graph NAS methods.

To address these shortcomings we propose a
probabilistic dual architecture search.
Instead of evaluating child networks from a parent network
iteratively using RL,
we train a superset of operations
with probabilistic priors
generated from a
NAS controller.
The controller then learns
the probabilistic distributions of candidate operators and picks
the most effective one from the superset.
For the macro-architecture,
we use the Gumbel-sigmoid trick
\cite{jang2016categorical, maddison2016concrete}
to relax discrete decisions
to be continuous,
so that a set of continuous variables can
represent the connections between graph blocks.
The proposed probabilistic, gradient-based NAS framework
optimises both the micro- and macro-architecture of GNNs.
Furthermore, we introduce several tricks
to improve both the search quality and speed.
First, we design the NAS controller to produce
multi-hot decision vectors to reduce
the combinatorial micro-architecture search dimensions.
Second, we use temperature annealing for Gumbel-sigmoid to
balance between exploration and convergence.
Third, our differentiable search is single-path, where
only a single operation from the superset is evaluated during each
training iteration This reduces the computation cost of NAS to the same as normal training.
In short, we make the following contributions in this paper:
\begin{itemize}
    \vspace{-3pt}
    \item
    We propose the first probabilistic dual network architecture
    search (PDNAS) method for GNNs.
    The proposed method uses Gumbel-sigmoid to relax
    the discrete architectural decision to be continuous
    for the macro-architecture search.
    \vspace{-3pt}
    \item
    To our knowledge, this is the first NAS that explores
    the macro-architecture space. We demonstrate how this
    helps deeper GNNs to achieve state-of-the-art results.
    \vspace{-3pt}
    \item We show several tricks (multi-hot controller, temperature annealling
    and single-path search) to improve the NAS search speed and quality.
    \vspace{-3pt}
    \item We present the performance of the networks
    discovered by PDNAS and show that they achieve
    superior accuracy and F1 scores
    in comparison to other hand-designed and NAS-generated networks
\end{itemize}

\section{Background}
\subsection{Network Architecture Search (NAS)}
DNNs achieve state-of-the-art results on a wide range of tasks,
but tuning the architectures of DNNs
on custom datasets is increasingly difficult.
One challenge is the increase in the number of different possible
operations that may be employed,
\eg~in the field of computer vision,
simple convolutions and fully connected layers
\cite{krizhevsky2012imagenet}
have expanded to
include depth-wise separable convolutions \cite{howard2017mobilenets},
grouped convolutions \cite{zhang2017interleaved},
dilated convolutions \cite{yu2017dilated}, \etc.
This opens up a much larger design space for
neural network architectures.
Network Architecture Search (NAS) seeks to automate this search for
the best DNN architecture. Initially NAS methods employed
reinforcement-learning (RL)
\cite{zoph2016neural,tan2019mnasnet}.
A recurrent neural network acts as a controller
and maximises the expected accuracy of the search
target on the validation dataset.
However, each update of the controller requires
a few hours to train a child network to convergence
which significantly increases the search time.
Alternatively, \citet{liu2018darts} proposed
Differentiable Architecture Search (DARTS)
that is a purely gradient-based search;
each candidate operation's importance
is scored using a trainable scalar and updated
using Stochastic Gradient Descent (SGD).
Subsequently, \citet{casale2019probabilistic}
approached the NAS problem from a probabilistic view,
transforming concrete trainable scalars used by DARTS \cite{liu2018darts}
to probabilistic priors and
only train a few architectures sampled from these priors
at each training iteration.
\citet{wu2019fbnet} and \citet{xie2018snas}
used the Gumbel-softmax trick
to relax discrete operation selection to continuous random
variables.
Existing NAS methods focus mainly on finding
optimal operation choices
inside each candidate block (micro-architecture),
in our work,
we extend the search to
consider how blocks are interconnected, \ie~
the network's macro architecture.

\subsection{NAS for GNNs}
While NAS methods have been developed using image and sequence data,
few recent work has applied them to graph-structured data.
\citet{gao2019graphnas}
first proposed GraphNAS,
a RL-based NAS on graph data.
\citet{zhou2019auto} used a similar RL-based approach (AutoGNN)
with a constrained parameter sharing strategy.
However, both of these NAS methods for graphs focus solely on
the micro-architecture space --- they search only which operations
to apply on individual graph blocks and do not learn
how large graph blocks connect to each other.
Moreover, these methods are RL-based;
to fully train the RL controller,
they require many iterations
of child network training to convergence.

In this work we focus on GNNs
applied to node classification tasks
based on Message-Passing Neural Networks
\cite{gilmer2017neural}.
Most of the manually designed architectures
proposed for these tasks fall into this category,
such as GCN~\cite{kipf2016semi},
GAT~\cite{velickovic2018graph},
LGCN~\cite{gao2018large} and GraphSage~\cite{hamilton2017inductive}.

\section{Method}
\begin{figure*}[!ht]
	\begin{center}
		\begin{subfigure}[b]{\linewidth}
		\centering
        \includegraphics[width=0.85\linewidth]{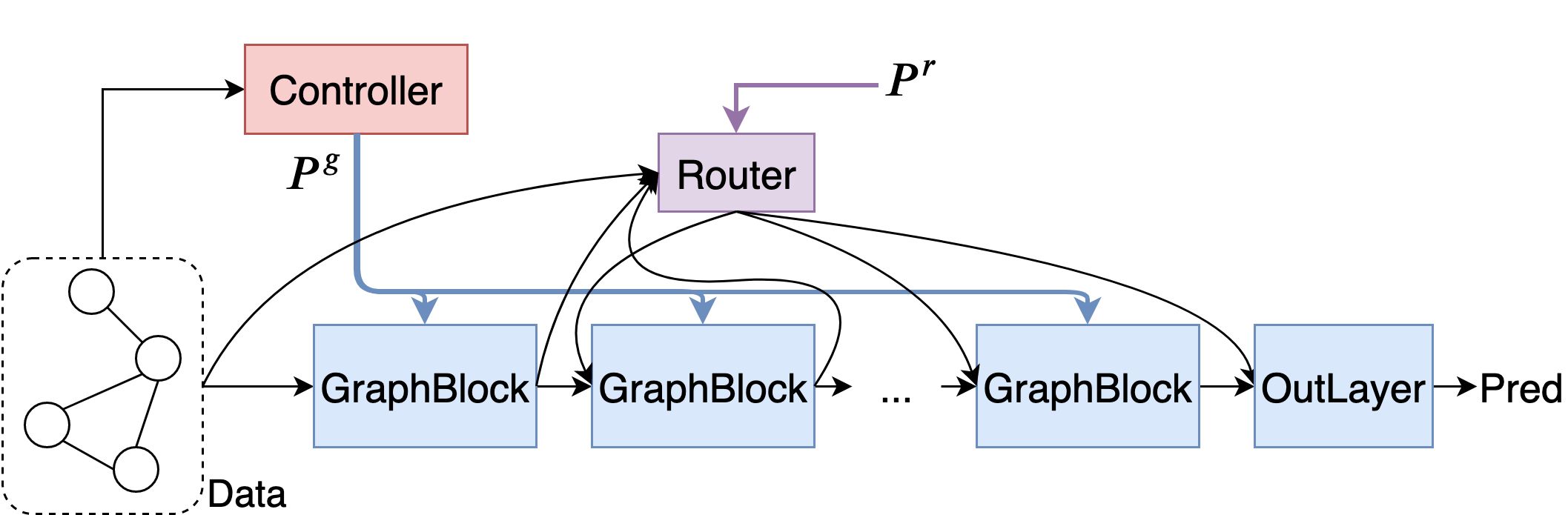}
        \caption{\label{fig:overview}}
        \end{subfigure}
		\begin{subfigure}[b]{0.64\linewidth}
			\includegraphics[width=\linewidth]{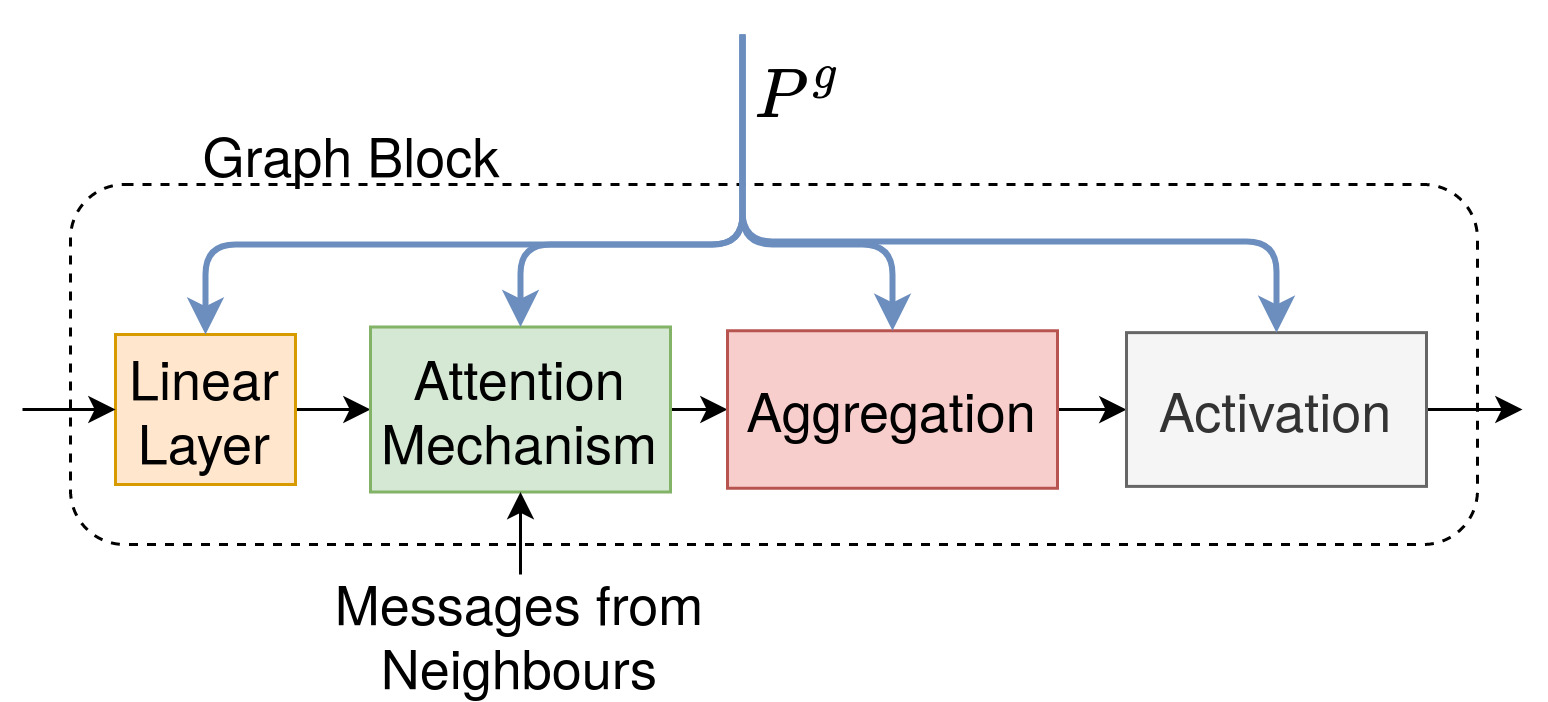}
			\caption{\label{fig:gb}}
		\end{subfigure}
		\begin{subfigure}[b]{0.34\linewidth}
			\includegraphics[width=\linewidth]{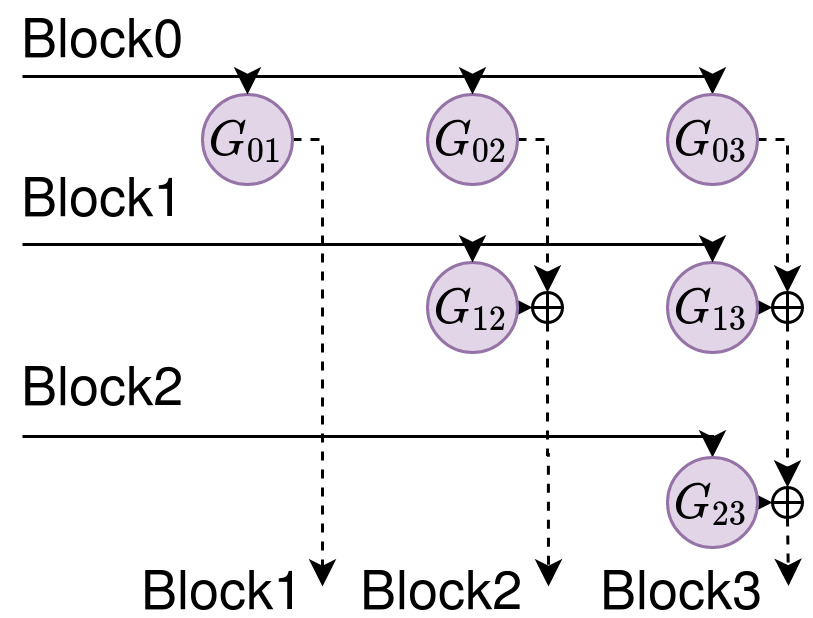}
			\caption{\label{fig:router}}
		\end{subfigure}
	\end{center}
	\caption{(a) Overview of PDNAS. $P^g$ denotes controller-output probabilities for operators within each graph block. $P^r$ denotes probabilities of shortcut connections between graph blocks. (b) Within a graph block, $P^g$ controls which operators are used for different types of operations. (c) $G_{ij}$ are gating functions $G_{ij}(I_i) = I_i \times P^r_{ij}$. Solid lines are input streams (denoted as $I$) into the router while dashed lines are output streams (denoted as $O$). $O_j = \sum_i G_{ij}(I_i)$.   }
	\label{fig:arch}
\end{figure*}

Figure~\ref{fig:overview} shows an overview of PDNAS.
In this framework,
we formulate the search space for GNN
as a stack of Graph Blocks (\Cref{fig:overview}),
with shortcut connections allowing information
to skip an arbitrary number of blocks, similar to DenseNet~\cite{huang2017densely}.
A Graph Block is essentially a GNN layer composed from four
sub-blocks, including a linear layer, an attention layer,
an aggregation and an activation function (\Cref{fig:gb}).
Each sub-block has a set of candidate operations to search over.
A NAS controller determines which operators are active
in these sub-blocks
during each training iteration.
While searching in the micro-architecture space of operations within Graph Blocks,
PDNAS also searches in the macro-architecture space of shortcut connections(\Cref{fig:router}).
Shortcut connections are controlled by gating functions
parameterised by a routing probability matrix.
We discuss the micro-architecture search of
a Graph Block in \Cref{sec:search_space},
the macro-architecture search of shortcut connection routing in \Cref{sec:routing}
and dual optimisation of architectural parameters in \Cref{sec:optimisation}.

\subsection{Micro-Architecture Search}\label{sec:search_space}
In this work, we consider GNNs based on the message-passing mechanism.
In each GNN layer,
nodes aggregate attention weighted messages from their neighbours
and combine these messages with their own feature.
Formally, each GNN layer can be described as:
\begin{equation}
\begin{split}
e^k_i &= \mathsf{AGGREGATE}_{j \in N(i)}(a^k_{ij} F^k(h^{k-1}_j)), \\
h^k_i &= \sigma(\mathsf{COMBINE}(e^k_i,F^k(h^{k-1}_i))).
\end{split}
\end{equation}
Here $F^k$ is a transformation operation for features.
In GNNs, $F^k(x)$ is typically a linear transformation in the form of $W^k x$.
$N(i)$ is the set of neighbouring nodes of node $i$.
$a^k_{ij}$ are the attention parameters for messages passed from neighbouring nodes.
$\mathsf{AGGREGATE}$ is an aggregation operation for the messages received.
$\mathsf{COMBINE}$ is an operation for combining aggregated messages
with features of the current node.
$\sigma$ is a non-linear activation.
For each of the above operations, there are several candidates to search amongst.
In this work, we consider the following micro-architecture search space:
\begin{table}[!h]
\caption{Different types of attention mechanisms. $W$ here is parameter vector for attention. $<,>$ is dot product, $a_{ij}$ is attention for message from node $j$ to node $i$.}
\label{tab:attn_types}
\vskip 0.15in
\begin{center}
\begin{small}
\begin{sc}
\begin{tabular}{c|l}
\toprule
Attention Type & Equation\\
\midrule
Const & $a_{ij} = 1$ \\
GCN & $a_{ij} = \frac{1}{\sqrt{d_i d_j}}$\\
GAT & $a^{gat}_{ij} = \mathsf{LeakyReLU}(W_a (h_i||h_j))$ \\
Sym-GAT & $a_{ij} = a^{gat}_{ij} + a^{gat}_{ji}$\\
COS & $a_{ij} = <W_{a1} h_i, W_{a2} h_j> $\\
Linear & $a_{ij} = \mathsf{tanh}(\sum_{j \in N(i)}(W_a h_j)) $\\
Gene-Linear & $a_{ij} = W_g \mathsf{tanh}(W_{a1} h_i + W_{a2} h_j) $ \\
\bottomrule
\end{tabular}
\end{sc}
\end{small}
\end{center}
\vskip -0.1in
\end{table}

\begin{itemize}
	\item \emph{Transformation function}:
	we formulate $F^k(x)$ as $W^k_2 \sigma(W^k_1 x)$
	where $W^k_1 \in \realset^{D_E\times D_I}$ and
	$W^k_2 \in \realset^{D_O\times D_E}$.
	$D_I$ and $D_O$ are the input and output dimensions for $F^k$ respectively,
	and $D_E$ is the expansion dimension,
	which is similar to \citet{tan2019mnasnet}.
	We let $D_E$ be multiples of $D_I$.
	The search space for $D_E$ is thus $\{D_I,2D_I,4D_I,8D_I\}$.
	While it is possible to search for the output dimension of $F^k$,
	this incurs large memory costs because a quadratic number of
	candidate operators are needed for each layer.
	Let the number of candidate hidden dimensions be $M$. For each layer we will have to use $M\times M$ candidate operators
	for each layer to map $M$ input dimensions to $M$ output dimensions.
	In our setup, we thus
	leave input dimensions
	as hyper-parameters determined via a grid search.

	\item \emph{Attention mechanism}:
	attention parameter $a^k_{ij}$ is computed by attention functions
	that may depend on the features of the current node and its neighbours.
	While any attention function can be included in the set of candidate operators,
	we use attention functions that appear
	in GraphNAS~\cite{gao2019graphnas} and AGNN~\cite{zhou2019auto}
	to make a fair comparison.
	\Cref{tab:attn_types} lists all attention functions
	considered in our search.

	\item \emph{Attention head}:
	multi-head attention~\cite{vaswani2017attention}
	means that
	multiple attention heads are used and computed in parallel.
	For PDNAS, the numbers of heads searched are $\{1,2,4,8,16\}$.

	\item \emph{Aggregation function}:
	messages from neighbouring node are aggregated
	by the function $\mathsf{AGGREGATE}$.
	In our experiments, we include three widely-used options:
	$\{\mathsf{SUM}, \mathsf{MEAN}, \mathsf{MAX\_POOLING\}}$.

	\item \emph{Combine function}:
	aggregated neighbouring messages are combined into
	the current node feature with the function $\mathsf{COMBINE}$.
	We examined two options,
	${\mathsf{ADD}}$ and ${\mathsf{CONCAT\_MLP}}$.
	Here, $\mathsf{ADD}$ is simply the addition of aggregated messages from
	neighbouring nodes to current node feature,
	while $\mathsf{CONCAT\_MLP}$ concatenates aggregated messages
	with node feature and then processes the result with a multi-layer perceptron (MLP).
	In practice,
	we found that $\mathsf{ADD}$ consistently outperforms the other,
	and thus removed the search for the combine function
	in our final implementation.

	\item \emph{Activation function}:
	the final output of a GNN layer passes
	through a non-linear activation function $\sigma$.
	The candidate functions for $\sigma$ include
	``None'', ``Sigmoid'', ``Tanh'', ``Softplus'',
	``ReLU'', ``LeakyReLU'', ``ReLU6'' and ``ELU''.
	Please refer to Appendix A 
	for details of each activation function.
\end{itemize}

A Graph Block is similar to a cell employed in the CNN NAS algorithm DARTS~\cite{liu2018darts}.
DARTS uses a weighted sum to combine
outputs of all candidate operators.
In PDNAS, we use the $\argmax$ function,
which allows only one candidate operator to be active in each training iteration.
Let $\bar{o}_{i,k}$ be the $k^{th}$ sub-block in Graph Block $i$,
and $o_{i,k,t}$ be the $t^{th}$ candidate operator for $\bar{o}_{i,k}$. $\bar{o}_{i,k}$ is then computed as:
\begin{equation}\label{eq:op_select}
\bar{o}_{i,k} = o_{i,k,t^{max}_{i,k}},
\text{where~} t^{max}_{i,k}
= \argmax_{t\in T} P^{g}_{i,k,t}.
\end{equation}
Here, $P^{g}_{i,k,t}$ is the probability of the $t^{th}$ candidate operator
of sub-block $k$ and layer $i$ assigned by the NAS Controller.
This hard-max approach considerably reduces memory and computational cost
since only one operation is active at any training iterations,
whilst still converges,
as shown by \citet{wu2019fbnet} and \citet{xie2018snas}.
While the $\argmax$ function is non-differentiable,
we use a differentiable approximation
which ensures that the controller receives learning signals.
The operator selection is implemented by casting $t^{max}_{i,k}$
as a one-hot vector $V_{i,k}$
to select from the outputs of each candidate operators.
We multiply this vector ($V_{i,k}$) with $P^{g}_{i,k}$
to allow gradients to be back-propagated
through $P^g_{i,k}$ to the controller.
This is the same as adding winner-takes-all
to the softmax-weighted summation
used in DARTS~\cite{liu2018darts},
and also known as single-path NAS.
In a single-path NAS,
only the winning operation is evaluated
during each training iteration,
the forward and backward passes
through the unselected operators are thus not evaluated.
It in turn reduces
the computational and memory costs of each iteration
to the same as normal training.
\begin{figure}[h]
	\begin{center}
        \includegraphics[width=1\linewidth]{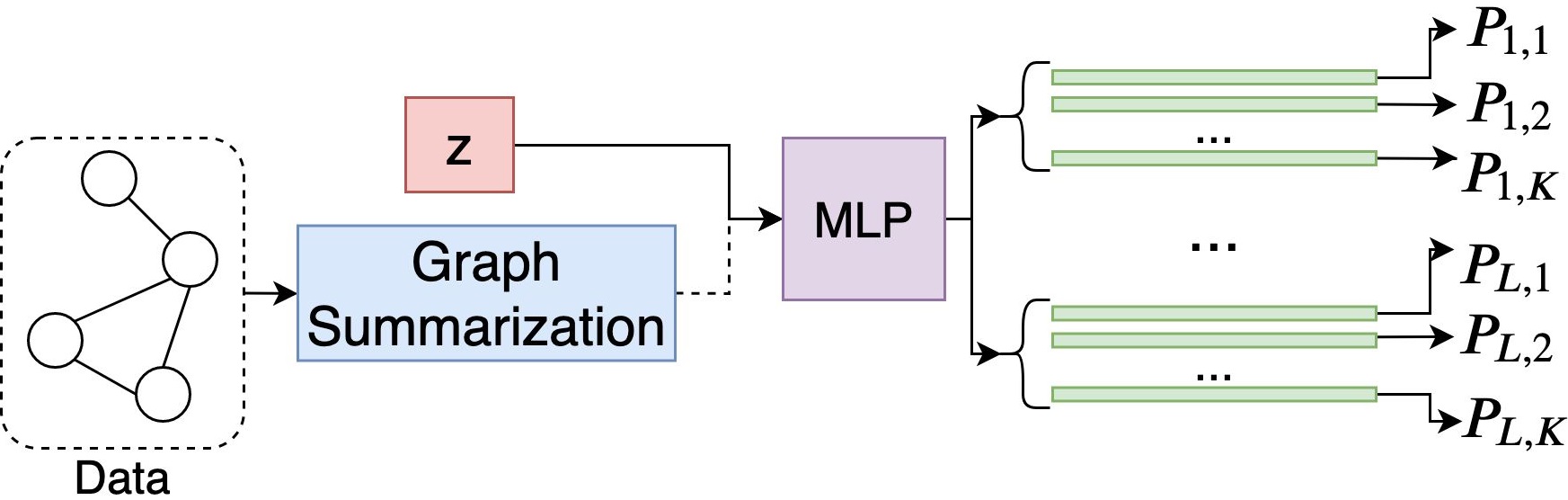}
	\end{center}
	\caption{
		Micro-architecture search controller overview.
		Here $z$ is the trainable prior, ``MLP" means Multi-Layer Perceptron.
		$P_{i,k}$ is a probability vector for operation $k$ in layer $i$.
		Dashed line means an optional path. }
	\label{fig:controller}
\end{figure}

\textbf{NAS Controller}:
\Cref{fig:controller} illustrates
the design of our micro-architecture search controller.
The controller is conditioned on two possible inputs,
which are a trainable prior vector $z$ and
a graph embedding $B_g$ produced by the graph summarisation module.
Graph summarisation module, as its name suggests,
summarises the whole graph into
a single vector embedding containing the entire dataset statistics.
In this work, we use a simple module with two GCN layers~\cite{kipf2016semi}
and two pooling layers after each GCN.
The first pooling layer is a self-attention pooling layer~\cite{lee2019self}
while the last layer is a global average pooling.
The graph summarisation module allows the NAS controller
to be conditioned on  input data.
We found that the performance improvement
provided by a graph summarisation path in the controller
is minimal ($<0.2\%$),
but it caused considerable additional computational and memory costs.
We then make this conditioning on data optional,
and report experiment results without this branch of data conditioning.

We combine $z$ and $B_g$ by $W_B B_g + z$,
thereby treating $z$ as a trainable bias to the data statistics.
The final part of the NAS controller is an MLP,
which computes $L\times K$ vectors
for every $K$ possible sub-block in each of the $L$ layers.
Each vector is passed through a softmax function to
produce a probability vector $\bar{P}^g_{i,k}$ that controls
which operator is active with the $\mathsf{argmax}$ function described in equation~\ref{eq:op_select}.
This multi-hot vector approach reduces parameters of the controller
considerably compared to the one-hot vector approach
where each whole-architecture configuration is represented
as a separate entry in the output vector.
The one-hot approach
will have an output layer of size complexity $O(T^{LK})$,
whereas our multi-hot approach
only requires an output layer of size $O(LKT)$%
\footnote{%
	For clarity,
	we assume the number of candidate operators $T$
	remain the same across sub-blocks in each layer.
}.

In initial experiments,
we found that when selecting the attention mechanism,
the NAS controller usually converges to operators that do not have trainable parameters,
such as GCN's normalised message weighting and constant attention.
We hypothesise this is because operators with trainable parameters,
such as GAT,
take many training iterations to achieve similar performance to parameter-less operators like GCN.
The controller thus at the start of training greedily converge to these parameter-less
operators due to a faster improvement in performance.
To enforce more ``exploration'',
we add noise to the probability distribution for operators generated by our NAS controller
at the start of the searching,
and gradually anneal the noise to 0.
Specifically the noise-added probability vector $P^g_{i,k}$ is computed as:
\begin{equation}\label{eq:noise}
P^g_{i,k} = \frac{\bar{P}^g_{i,k} + \tau U(0,1)}{Z}.
\end{equation}
Here, $U$ is a uniform distribution to sample noise from,
$\tau$ is the temperature which decreases during search to anneal noise,
and $Z$ is a normalising factor to ensure $P_{i,k}$ is still a valid probability distribution.
While it is possible to use
Gumbel-Softmax~\cite{jang2016categorical,maddison2016concrete}
to achieve the same goal. In practice, we found that the controller
greedily increase the scale of logits inputs,
making the Gumbel noise too small to make any effect.
Thus we enforce inputs and the noise to be at the same numerical scale using \Cref{eq:noise}.

\subsection{Macro-architecture Search}\label{sec:routing}

The macro-architecture search determines
how graph blocks connect to each other,
in this case,
we call them \emph{shortcut connections}
following the naming conventions in computer vision \cite{huang2017densely}.
As mentioned earlier,
shortcut connections on graph data have been explored
in Jumping Knowledge networks \cite{xu2018representation}.

We define
$\mymatrix{\bar{P}^r} \in \realset^{L \times L}$
to be a square matrix of trainable priors
for shortcut connections,
and $L$ is the number of possible graph blocks.
Additionally,
$\mymatrix{P^r}$ denotes a collection of the probabilities of
connection between the inputs and outputs
of graph blocks through shortcut connections,
and has the same dimension as $\mymatrix{\bar{P}^r}$.
In addition, cyclic connections are not permitted.
$\mymatrix{I}$ is a collection of $L$ inputs,
where $I_i$ represents a single graph input from
a previous layer and $ 0 \leq i \leq L$.
Similarly, $\mymatrix{O'}$ is a collection of $L$ output graphs;
these are the original outputs of graph blocks.
With $O_j'$ being a single graph output,
we have $ 0 \leq j \leq L$;
$\mymatrix{O}$ has the same dimension as $\mymatrix{O'}$,
and it is the combination between shortcut connections
and the original outputs.
For producing the probabilities $\mymatrix{P^r}$ of shortcut connections
from trainable priors $\mymatrix{\bar{P}^r}$,
we apply the Gumbel-Sigmoid trick
\cite{jang2016categorical,maddison2016concrete}
(denoted as $\mathsf{gs}$)
on each individual element of $\mymatrix{P^r}$
so as to approximate discrete sampling
from a binomial distribution.
Gumbel-Sigmoid has the form of:
\begin{equation}\label{eq:gumbel_sigmoid}
y = \frac{\mathsf{exp}(\frac{\mathsf{log} \, a +g }{\tau})}{1 + \mathsf{exp}(\frac{\mathsf{log} \, a+g}{\tau})},
\end{equation}
where $g$ is noise sampled
from the Gumbel distribution $\mathsf{Gumbel}(0,1)$,
and $\tau$ is the temperature controlling the randomness for the Gumbel statistics.
As $\tau$ decreases, $\mathsf{gs}$ samples values that are more `discrete',
meaning that values are closer to extreme boundary of 0 and 1,
$\mathsf{row\_sum}$ reduces the matrix by summing all row elements,
and $\boldsymbol{\odot}$ is the element-wise product
between matrices.
\begin{equation}
	\begin{split}
		\mymatrix{O} & = \mymatrix{O'} + G(\mymatrix{I}, \mymatrix{\bar{P}^r}, \tau) \\
		& =
		\mymatrix{O'} +
		\mathsf{row\_sum}(
		\mathsf{gs}(\mymatrix{\bar{P}^r}, \tau) \boldsymbol{\odot} G'(\mymatrix{I}))
		\\
		& =
		\mymatrix{O'} +
		\mathsf{row\_sum}(
		\mymatrix{P^r} \boldsymbol{\odot} G'(\mymatrix{I})).
	\end{split}
\end{equation}
Here, $G'$ is a collection of shortcut connections, which is simply
a fully connected layer that transforms the hidden unit size.
In addition $G'$ is an upper triangular matrix because shortcuts
are forward connections --- no graph blocks can connect backwards:
\begin{equation}
	G'(\mymatrix{I}) =
	\begin{bmatrix}
		g_{00}(I_0)  & g_{01}(I_1) 	& \dots & g_{0j}(I_j) \\
		0  			 & g_{11}(I_1)  & \dots & g_{1j}(I_j) \\
		\hdotsfor{4} \\
		0        	 & 0 			& \dots & g_{ij}(I_j)
	\end{bmatrix}.
\end{equation}
We also have the following probability matrix $\mymatrix{\bar{P}^r}$,
note that this is an upper triangular matrix
with each $\bar{P}^r_{ij}=0$ if $i > j$.
This means input $I_i$ cannot
connect back to preceding $O_j$:
\begin{equation}
		\mymatrix{\bar{P}^r}
		=
		\begin{bmatrix}
			\bar{P}^r_{00}       & \bar{P}^r_{01}  & \dots & \bar{P}^r_{0j} \\
			0       	   & \bar{P}^r_{11}  & \dots & \bar{P}^r_{1j} \\
			\hdotsfor{4} \\
			0       	   & 0 		   & \dots & \bar{P}^r_{ij}
		\end{bmatrix}.
\end{equation}
For the $\mathsf{gs}$ (Gumbel-sigmoid) function,
we anneal the temperature to balance between
random choices and concrete discrete decisions.
With $e$ being the number of training epochs,
$e_m$ being the maximum number of epochs,
$\alpha$ being a constant and
$e_s$ is the starting epoch;
we use the following annealing strategy,
where in practice, we set $\alpha=1.0$, $e_s=80$:  
\begin{equation}
\tau =
\begin{cases}
    1, & \text{if } e <  e_s\\
    \mathsf{exp}^{-\frac{\alpha}{e_m} (e - e_{s})} , & \text{otherwise}
\end{cases}
\end{equation}

\subsection{Dual Optimisation}\label{sec:optimisation}
We formulate PDNAS as a bi-level optimisation problem,
similar to DARTS \cite{liu2018darts}:
\begin{equation}
\begin{split}
\min_{a} & \, \mathcal{L}_{val} (w^\star(a),a) \\
s.t. & \, w^\star(a) = \argmin_{w} (\mathcal{L}_{train}(w,a))
\end{split}
\end{equation}
Here $w$ are the parameters of all candidate operators,
$w^\star(a)$ is the optimal parameters given $a$, where $a$ represents
parameters of the micro-architecture search controller $a_{\mathsf{micro}}$
and the trainable routing matrices $a_{\mathsf{macro}}$.
$L_{train}$  is a training loss on the training data split,
while $L_{val}$ is validation loss on the validation data split.
The parameters $w$ and $a$ are trained iteratively
with their own gradient descent optimisers.
Since it is computationally intractable to compute $w^\star(a)$
for each update of $a$,
we approximate $w^\star$ with a few training steps,
which are shown to be effective in DARTS~\cite{liu2018darts},
gradient-based hyper-parameter tuning~\cite{luketina2016scalable}
and unrolled Generative Adversarial Network training~\cite{metz2016unrolled}.
The full procedure is shown in Algorithm~\ref{alg:opt}.
Here $x$ is input data, $y$ is label,
$\mathsf{MaxIter}$ is the maximum number of search iterations,
and $\mathsf{TrainStep}$ is the number of training steps to approximate $w^\star$.
In each search iteration,
we first sample noise $N$ for the controller
(recall this noise is to encourage more exploration at the start of training),
and then compute probabilities of the candidate operators $P^g$
and indices of operators with the highest probabilities $\mathsf{Index}$.
We then approximate $w^\star$ in $\mathsf{TrainStep}$ steps.
In the training steps,
$w$ of operators receives gradients from the optimiser $\mathsf{Opt}_w$
using the training loss $\mathcal{L}_{train}$.
Next we update both sets of architectural parameters
(controller and router parameters),
$a_{\mathsf{micro}}$ and $a_{\mathsf{macro}}$,
with respect to the validation loss $\mathcal{L}_{val}$.
Note here $\mathsf{Index}$ is changed to $P^g[\mathsf{Index}]$
to provide gradients to the controller,
as discussed in \Cref{sec:search_space}.
In practice we use the Adam optimiser~\cite{kingma2014adam},
noted as $\mathsf{Opt}$.
\begin{algorithm}[tb]
	\caption{Dual Architecture Optimisation}
	\label{alg:opt}
	\begin{algorithmic}
		\STATE {\bfseries Input:} $x_{\mathsf{train}}$, $y_{\mathsf{train}}$, $x_{\mathsf{val}}$, $y_{\mathsf{val}}$, $\mathsf{MaxIter}$, $\mathsf{TrainStep}$
		\STATE $\mathsf{Init}(w,a_{\mathsf{micro}}, a_{\mathsf{macro}},P^r)$

		\FOR{$e=0$ {\bfseries to} $\mathsf{MaxIter} - 1$}
		    \STATE $\tau$ = TempAnneal($e$)
		    \STATE $N$ = SampleNoise($\tau$)
		    \STATE $P^g$ = Controller($x_{\mathsf{val}}$,$N$)
		    \STATE $\mathsf{Index}_{i,k} = \argmax_{t}(P^g_{i,k,t})$
		    \FOR{$i=0$ {\bfseries to} $\mathsf{TrainStep} - 1$}
               \STATE $\mathcal{L}_{train} = \mathsf{Loss}(x_{\mathsf{train}}, y_{\mathsf{train}},\mathsf{Index},P^r)$
               \STATE $w$ = $\mathsf{Opt}_w (\mathcal{L}_{train})$
		    \ENDFOR
               \STATE $\mathcal{L}_{val} = \mathsf{Loss}(x_{\mathsf{val}}, y_{\mathsf{val}},P^g[\mathsf{Index}],P^r)$
               \STATE $a_{\mathsf{micro}}$ = $\mathsf{Opt}_{{\mathsf{micro}}} (\mathcal{L}_{val})$
               \STATE $a_{\mathsf{macro}}$ = $\mathsf{Opt}_{{\mathsf{macro}}} (\mathcal{L}_{val})$
		\ENDFOR
	\end{algorithmic}
\end{algorithm}

\section{Results}
We implemented PDNAS using PyTorch \cite{paszke2019pytorch}.
Operations in Graph Blocks are modified from the GNN implementations
in PyTorch Geometric (PyG) \cite{fey2019fast}.
For the Cora dataset, to ensure a consistent comparison
to GraphNAS \cite{gao2019graphnas},
we used the data splits provided by the Deep Graph Library
\cite{wang2019dgl}.
The data splits from all other datasets are from PyG.
For all search and training,
we used a single Nvidia Tesla V100 GPU unless specified otherwise.
We evaluated PDNAS on two learning settings,
namely transductive and inductive settings.
For the transductive setting we used
the citation graph datasets~\cite{sen2008collective} including Cora, Citeseer and PubMed.
For the inductive setting, we considered
the Protein-Protein Interaction (PPI) dataset~\cite{zitnik2017predicting}.
In addition, we provide an evaluation of the citation datasets
in a fully supervised setting, similar to \citet{xu2018representation}.

\subsection{Citation Datasets}
For Citation datasets,
we conducted the experiment with two widely-used settings with the
former according to \citet{yang2016revisiting}
and the latter according to \citet{xu2018representation}.
In this section, we describe the results for both settings.

In the first setting,
training data only contains 20 labelled nodes for each category in the dataset.
Validation data contains 500 nodes, while test data contains 1000 nodes.
We used a learning rate of $0.005$ for model parameters $w$ and $0.002$
for architectural parameters $a$, and ran search for 400 epochs.
In \Cref{tab:citation:original},
we present the results of PDNAS for this setting in comparison to
graph attention networks (GAT) \cite{velickovic2018graph},
GraphNAS \cite{gao2019graphnas} and AGNN \cite{zhou2019auto}.
The results demonstrate that PDNAS outperforms all existing
methods on Cora and PubMed, however, is $0.3\%$ lower
on Citeseer compared to AGNN.
In addition to accuracy,
we also measured the search wall clock times of GraphNAS~\cite{gao2019graphnas}
using their open sourced code \footnote{\url{https://github.com/GraphNAS/GraphNAS}}.
Unfortunately AGNN~\cite{zhou2019auto} does not have an open source implementation,
nor reports wall clock times, making it impossible to compare against.
Table~\ref{tab:time} shows wall clock times used for searching with GraphNAS and PDNAS.
The comparison is conducted with exactly the same software and hardware environments.
Time used for PDNAS takes into account of hyper-parameter search of hidden layer sizes,
as discussed in Section~\ref{sec:search_space}.
We see that PDNAS is more than two times faster than GraphNAS for finding the best GNN architecture.
\begin{table}[h!]
\caption{
    Accuracy comparison on Cora, Pubmed and Citeseer with data splits same as \citet{yang2016revisiting}.
    Our results are averaged across 3 independent runs.
    The numbers in bold show best accuracies.
}
\label{tab:citation:original}
\vskip 0.15in
\begin{center}
\begin{small}
\begin{sc}
\begin{tabular}{c|ccc}
\toprule
Methods   & Cora      & CiteSeer    & PubMed \\
\midrule
GAT & $83.0 \pm 0.7\%$  & $72.5 \pm 0.7 \%$ & $79.0 \pm 0.3\%$ \\
GraphNAS & $84.2 \pm 1.0 \%$ & $73.1 \pm 0.9 \%$ & $79.6 \pm 0.4 \%$ \\
AGNN & $83.6 \pm 0.3\%$  & $\mathbf{73.8 \pm 0.7 \%}$ & $\mathbf{79.7 \pm 0.4 \%}$ \\
\midrule
PDNAS & $\mathbf{84.5 \pm 0.6 \%}$ & $73.5 \pm 0.3 \%$ & $\mathbf{79.7 \pm 0.6 \%}$\\
\bottomrule
\end{tabular}
\end{sc}
\end{small}
\end{center}
\vskip -0.1in
\end{table}

\begin{table}[h!]
\caption{
    Comparison of wall clock time (measured in seconds) used on Cora, Pubmed and Citeseer with GraphNAS~\cite{gao2019graphnas}.
}
\label{tab:time}
\vskip 0.15in
\begin{center}
\begin{small}
\begin{sc}
\begin{tabular}{c|ccc}
\toprule
Methods   & Cora      & CiteSeer    & PubMed \\
\midrule
GraphNAS  & 11323 & 16333 & 26174\\
\midrule
PDNAS  & 5012 & 6044 & 10634\\
\bottomrule
\end{tabular}
\end{sc}
\end{small}
\end{center}
\vskip -0.1in
\end{table}

\begin{table*}[!h]
\caption{
    Accuracy and size comparison on Cora, Pubmed and Citeseer,
    the data split is $60\%$ training, $20\%$ validation and $20\%$ testing.
    JKNet-n is our implementation of a jumping knowledge network with concatenation as
    shortcut aggregation, n represents the channel count for each layer
    of the network.
    The numbers in bold are best accuracies for each model on the targeting datasets,
    numbers in shades are the best on each dataset across models.
    All accuracies are reported as averaged values from 3 independent runs.
    }
\label{tab:citation:partition}
\vskip 0.15in
\begin{center}
\begin{small}
\begin{sc}
\begin{tabular}{cc|cccccc}
\toprule
\multirow{2}*{Model} &
\multirow{2}*{Layers}
& \multicolumn{2}{c}{Cora} & \multicolumn{2}{c}{PubMed} & \multicolumn{2}{c}{Citeseer} \\
& & Accuracy & Size & Accuracy & Size  & Accuracy & Size\\
\midrule
\multirow{6}*{JKNet-32}
            & 2     & $\mathbf{89.28 \pm 0.00}$ & 48.30K & $\mathbf{88.54 \pm 0.03}$ & 17.67K & $\mathbf{75.49 \pm 0.00}$ & 120.74K \\
            & 3     & $88.35 \pm 0.00$ & 49.35K & $87.89 \pm 0.02$ & 18.72K & $73.68 \pm 0.00$ & 121.80K \\
            & 4     & $87.99 \pm 0.00$ & 50.41K & $87.30 \pm 0.06$ & 19.78K & $72.63 \pm 0.00$ & 122.85K \\
            & 5     & $87.99 \pm 0.00$ & 51.46K & $86.99 \pm 0.05$ & 20.84K & $72.08 \pm 0.10$ & 123.91K \\
            & 6     & $87.92 \pm 0.12$ & 52.52K & $86.97 \pm 0.02$ & 21.89K & $72.78 \pm 0.00$ & 124.97K \\
            & 7     & $88.10 \pm 0.12$ & 53.58K & $86.86 \pm 0.08$ & 22.95K & $72.23 \pm 0.10$ & 126.02K \\
\midrule
\multirow{6}*{JKNet-64}
            & 2     & $\mathbf{89.28 \pm 0.00}$ & 98.63K & $\mathbf{88.65 \pm 0.03}$  & 37.38K & $\mathbf{75.34 \pm 0.00}$ & 243.53K \\
            & 3     & $88.17 \pm 0.00$ & 102.79K & $87.95 \pm 0.05$ & 41.54K & $74.13 \pm 0.00$ & 247.69K \\
            & 4     & $87.98 \pm 0.00$ & 106.95K & $87.40 \pm 0.03$ & 45.70K & $72.78 \pm 0.00$ & 251.85K \\
            & 5     & $87.98 \pm 0.00$ & 111.11K & $87.29 \pm 0.13$ & 49.86K & $72.18 \pm 0.00$ & 256.01K \\
            & 6     & $87.80 \pm 0.00$ & 115.27K & $87.18 \pm 0.09$ & 54.02K & $72.28 \pm 0.00$ & 260.17K \\
            & 7     & $88.11 \pm 0.12$ & 119.43K & $87.24 \pm 0.10$ & 58.18K & $71.73 \pm 0.00$ & 264.33K \\
\midrule
\multirow{6}*{PDNAS}
            & 2     & $89.34 \pm 0.12$ & 48.06K & $89.14 \pm 0.19$ & 18.21K & \cellcolor{gray!25} $\mathbf{76.29 \pm 0.25}$ & 119.65K \\
            & 3     & $89.34 \pm 0.12$ & 50.22K & $89.14 \pm 0.19$ & 20.35K & $75.54 \pm 0.25$ & 123.59K \\
            & 4     & \cellcolor{gray!25} $\mathbf{89.77 \pm 0.31}$ & 51.29K & \cellcolor{gray!25} $\mathbf{89.25 \pm 0.08}$ & 25.67K & $75.64 \pm 0.15$ & 125.00K \\
            & 5     & $89.53 \pm 0.31$ & 57.66K & $89.24 \pm 0.08$ & 29.58K & $75.99 \pm 0.85$ & 129.97K \\
            & 6     & $89.53 \pm 0.31$ & 61.93K & $89.24 \pm 0.08$ & 32.43K & $75.74 \pm 0.20$ & 131.40K\\
            & 7     & $89.65 \pm 0.37$ & 68.65K & $89.24 \pm 0.08$ & 42.31K & $75.54 \pm 0.20$ & 141.64K \\

\bottomrule
\end{tabular}
\end{sc}
\end{small}
\end{center}
\vskip -0.1in
\end{table*}

\begin{table*}[!h]
\caption{
    Accuracy and size comparison on PPI\@.
    The symbol $^\star$ denotes it is an implementation from \citet{zhou2019auto}.
    The numbers in bold are the best F1 score for all models on this dataset,
    all F1 scores are reported as averaged values from 3 independent runs.
}\label{tab:ppi}
\vskip 0.15in
\begin{center}
\begin{small}
\begin{sc}
\begin{tabular}{ccc|cc}
\toprule
Model/Method       & Type    & Layers   & F1 Score            & Size \\
\midrule
GAT$^\star$            & Hand-Designed     & 3     & $97.30 \pm 0.20$    & 0.89M \\
LGCN$^\star$       & Hand-Designed     & 2     & $77.20 \pm 0.20$    & 0.85M \\
JKNet-Concat \cite{xu2018representation} & Hand-Designed    & 2     & $95.90 \pm 0.30$    & - \\
JKNet-LSTM \cite{xu2018representation}  & Hand-Designed    & 3     & $96.90 \pm 0.60$    & - \\
JKNet-Dense-LSTM \cite{xu2018representation} & Hand-Designed & 3    & $97.60 \pm 0.70$    & - \\
\midrule
GraphNAS \cite{gao2019graphnas}    & Reinforcement Learning        & 3     & $98.60 \pm 0.10$    & 3.95M \\
GraphNAS with sc \cite{gao2019graphnas} & Reinforcement Learning   & 3     & $97.70 \pm 0.20$    & 2.11M \\
AGNN \cite{zhou2019auto}        & Reinforcement Learning        & 3     & $99.20 \pm 0.20$    & 4.60M \\
AGNN with sharing \cite{zhou2019auto} & Reinforcement Learning    & 3     & $99.10 \pm 0.10$    & 1.60M \\
PDNAS        & Gradient-Based    & 4     & $\mathbf{99.27 \pm 0.03}$    & 2.39M \\

\bottomrule
\end{tabular}
\end{sc}
\end{small}
\end{center}
\vskip -0.1in
\end{table*}

It is worth mentioning that the original data splits on the citation
datasets are not suitable for training deeper graph networks;
the number of available training nodes is significantly smaller
than both validation and testing.
In other words, the search for the best network architecture with
limited number of training samples becomes an optimisation
focusing on micro-architectures.
Deeper networks are not applicable on such datasets since
over-fitting occurs easily with a small number of training samples.

To overcome the issue of the original unfair data splits, in the second setting,
we randomly repartitioned the datasets into $60\%$, $20\%$, $20\%$
for training, validation and testing respectively.
The random partition remains the same for all different networks
examined in \Cref{tab:citation:partition}.
It is notable that \citet{xu2018representation}
also repartitioned their data to the same $60\%$, $20\%$, $20\%$ split,
however, due to the unavailability of their data split masks,
we chose to reimplement their networks on our own random split.
\Cref{tab:citation:partition} shows
a comparison between manually-designed
jumping knowledge networks (JKNets) \cite{xu2018representation} and
our searched networks on the citation network datasets
\cite{yang2016revisiting}
(Cora, Pubmed and Citeseer).
Since the original JKNet 
can have varying numbers of channels at each layer of the
network, we implemented two versions with 32 channels and 64 channels
for each layer of the network respectively.
For both our search method and JKNets, we sweep the number of layers
from $2$ to $7$.
For each accuracy number reported in \Cref{tab:citation:partition},
it is averaged across $3$ independent runs;
in addition, the standard deviation among $3$ runs is also reported.
In practice, for searched networks, the network sizes for multiple
independent runs only vary slightly
and thus are not shown here for the ease of presentation.
The results in \Cref{tab:citation:partition} suggest our searched
networks outperformed JKNet by a significant margin.
For the best performing configuration on each model,
we observed increases of $0.49\%$, $0.6\%$ and $0.95\%$
in the average
accuracy on Cora, Pubmed and Citeseer respectively
(numbers in bold).
For both Cora and Pubmed, the best performing searched networks
are at a higher layer count compared to JKNets, demonstrating
our search algorithm is efficient at finding deeper networks.

\subsection{PPI dataset}
\Cref{tab:ppi} shows a comparison among several
hand-designed networks and various NAS results on the PPI dataset
\cite{zitnik2017predicting}.
The networks include
Graph Attention Networks (GAT) \cite{velickovic2018graph},
learnable graph convolutional networks (LGCN) \cite{gao2018large},
and jumping knowledge networks (JKNet) \cite{xu2018representation}.
Jumping knowledge networks did not report the size and the
original code base is not available, so we do not report their
sizes.
For the network architecture search results, we compare
to GraphNAS \cite{gao2019graphnas} and AutoGNN \cite{zhou2019auto}.
Both of these NAS methods are RL-based
and do not support searching on a macro-architecture level.
As a result, our search method finds a deeper network with
the highest F1 score in comparison to the other NAS methods.
PDNAS outperforms the best hand-designed network and
NAS network by $1.67$ and $0.17$ respectively.

\section{Conclusion}

In this paper we provide evidence that a differentiable and
dual-architecture approach to NAS can outperform current NAS approaches applied to GNNs,
both in terms of speed and search quality.
The micro-architecture
design space is searched using a pure gradient-based approach and
search complexity is reduced using a multi-hot NAS controller.
In addition, for the first time, NAS is extended to consider the
network's macro-architecture using a differentiable routing mechanism.

\clearpage

\nocite{langley00}
\bibliography{references}
\bibliographystyle{icml2020}

\end{document}